\documentclass[letterpaper, 10 pt, conference]{ieeeconf}  

\IEEEoverridecommandlockouts                              

\usepackage{graphics} 
\usepackage{epsfig} 
\usepackage{mathptmx} 
\usepackage{times} 
\usepackage{amsmath} 
\usepackage{amssymb}  

\usepackage{url}
\usepackage{multirow}
\usepackage{graphicx}
\usepackage[table,xcdraw]{xcolor}
\usepackage{makecell}
\usepackage{boldline}

\usepackage{caption}
\usepackage{subcaption}
\usepackage{tabulary}

\usepackage{dsfont}
\usepackage{booktabs}

\usepackage{url}

\newcommand{\ie}{\textit{i}.\textit{e}.}

\title{\LARGE \bf
Robust Autonomous Landing of UAV in Non-Cooperative Environments based on Dynamic Time Camera-LiDAR Fusion
}

\author{Lyujie Chen$^{1}$, Xiaming Yuan$^{1}$, Yao Xiao$^{2}$, Yiding Zhang$^{3}$ and Jihong Zhu$^{2}$
\thanks{$^{1}$The authors are with the Department of Computer Science and Technology, Beijing National Research Center for Information Science and Technology, Tsinghua University, Beijing, 100084 (e-mail: {\tt\small jimchen0605@gmail.com}).}%
\thanks{$^{2}$The authors are with the Department of Precision Instrument, Tsinghua University, Beijing (e-mail: {\tt\small jhzhu@tsinghua.edu.cn}).}%
\thanks{$^{3}$The author is with the Department of Automation, Tsinghua University, Beijing, 100084.}%
}

\begin{document}

\maketitle
\thispagestyle{empty}
\pagestyle{empty}

\begin{abstract}

Selecting safe landing sites in non-cooperative environments is a key step towards the full autonomy of UAVs. However, the existing methods have the common problems of poor generalization ability and robustness. Their performance in unknown environments is significantly degraded and the error cannot be self-detected and corrected. In this paper, we construct a UAV system equipped with low-cost LiDAR and binocular cameras to realize autonomous landing in non-cooperative environments by detecting the flat and safe ground area. Taking advantage of the non-repetitive scanning and high FOV coverage characteristics of LiDAR, we come up with a dynamic time depth completion algorithm. In conjunction with the proposed self-evaluation method of the depth map, our model can dynamically select the LiDAR accumulation time at the inference phase to ensure an accurate prediction result. Based on the depth map, the high-level terrain information such as slope, roughness, and the size of the safe area are derived. We have conducted extensive autonomous landing experiments in a variety of familiar or completely unknown environments, verifying that our model can adaptively balance the accuracy and speed, and the UAV can robustly select a safe landing site.

\end{abstract}

\section{Introduction}

Autonomous landing is a crucial function of Unmanned Aerial Vehicles (UAVs) and the foundation of various civilian and military missions, such as search and rescue, payload delivery, and emergency landing. Due to the complexity of the near-ground situations, it is also the most accident-prone task. Therefore, the ability to select safe landing sites in complex and unknown environments is essential to improve the autonomy level of UAVs.
Most of the previous related works~\cite{sharp2001vision,saripalli2002vision,saripalli2003visually,miller2008landing,lange2009vision,lee2012autonomous,kong2013autonomous,kong2014ground,muskardin2016landing,benini2016real,wang2018hierarchical} conduct autonomous landing in the cooperative environment, that is, either directly exchange information through wireless communication~\cite{kong2013autonomous,kong2014ground}, or there is a pre-defined landing mark or landing places in the scene, such as graphical signs~\cite{sharp2001vision,saripalli2003visually,lange2009vision,benini2016real}, two-dimensional bar code signs~\cite{lee2012autonomous,muskardin2016landing,wang2018hierarchical}, or natural runways~\cite{saripalli2002vision,miller2008landing}. In this way, because the characteristics of the landing locations are known in advance, these works more focus on calculating the relative position and attitude between the UAV and landing site.

However, in the non-cooperative environment, the major problem is to select a safe landing site without guidance where various factors should be considered, including detecting the flatness and roughness of the ground, distinguishing terrain types, calculating landing area size, and avoiding obstacles. 
Previous works propose several approaches to solve it, such as depth/altitude estimation, surface type semantic segmentation, and high-precision map construction. Specifically, \cite{papa2015design} uses ultrasonic sensors to perceive depth for UAV safe landing at a low-altitude (150cm) range. \cite{eynard2010uav,park2014landing} estimate the altitude in indoor scenes with the widely-used stereo vision system. \cite{johnson2005vision,cherian2009autonomous,gordon2019depth} apply traditional or machine-learning-based methods to realize monocular depth prediction. \cite{guo2014automatic,guo2016robust} perform patch-level classification to evaluate the safeness of the ground. \cite{bosch2006autonomous} uses the combination of a robust homography estimation and a strategy of adaptive thresholding of correlation scores to realize a pixel-level segmentation. \cite{patterson2011utilizing} utilizes geographic information system data to estimate the landing position. Some approaches~\cite{hinzmann2018free,park2014landing} combine multiple criteria, such as altitude, flatness, roughness, slope, surface type, and energy consumption to jointly determine the landing sites. To a step further, ~\cite{marcu2018safeuav} directly detects the plane of the ground in an end-to-end way learning with a virtual image dataset. However, the learning target is too high-level, leading to an over-fitting result and poor generalization.

Among these methods, depth/altitude estimation is most feasible since it is the basis for calculating the terrain characteristics and is also a widely studied area. Benefiting from the advantages of low cost, lightweight, and long viewing range, vision-based depth estimation is a common approach, including monocular depth prediction and stereo depth matching.
Currently, deep-learning-based depth estimation methods~\cite{eigen2014depth, laina2016deeper, godard2017unsupervised, fu2018deep} are prevalent due to their high accuracy. However, applying it to the aerial scenes is difficult and rarely used. One reason is that although there are some virtual aerial datasets~\cite{Fonder2019MidAir,chen2020valid} containing annotated depth maps, it still lacks reliable large-scale real scenes training data. Another reason is that the current deep-learning-based methods do not have reliable generalization ability, resulting in a great degradation of performance in unknown environments. What's more, the prediction errors cannot be self-detected under their end-to-end model strategies. For this reason, a feasible solution is to integrate a reliable depth sensor, such as ultrasonic wave sensors, Millimeter Wave Radar (MMWR), and Light Detection And Ranging (LiDAR), to formulate a depth completion problem with partial known depth. Among these sensors, LiDAR is widely used in unmanned driving and robotics to integrate with vision systems due to its long detection range and high accuracy. But the current mainstream products are too expensive and heavy, thus are rarely used in UAV systems.

In this paper, we build a UAV platform equipped with a low cost and lightweight binocular-LiDAR system to realize robust and safe autonomous landing. Our LiDAR (Livox Mid-40) uses a single line to scan towards one side in a non-repetitive scanning pattern in which the areas scanned inside the Field of View (FOV) grow the longer the integration time until the FOV coverage approaches 100\%. Using this feature, we propose a dynamic time depth completion algorithm. Specifically, in the training phase, the stereo camera and LiDAR provide the self-supervision and sparse ground truth respectively for joint training. While in the inference phase, we introduce a self-evaluation method to assess the accuracy of the predicted depth map using the character of the binocular camera. It realizes the dynamic selection of the LiDAR accumulation time and ensures the high quality of the predicted depth map, which greatly improves the generalization ability of the model. With the high FOV coverage advantage, even in the extreme unknown environment, accurate depth measurement can still be obtained through the LiDAR, which enhances the overall safety and reliability at the expense of reducing the speed of landing. Based on the estimated depth map, the first and second-order derivatives stand for the slope and roughness of the ground respectively. Then, the flatness and the safe area size can be easily derived.

We collect 10 real flight records in different environments using onboard sensors. By performing spatial and time alignment with high-frequency GPS and IMU data, 30,000 sparsely annotated depth maps are generated. On this dataset, our dynamic time depth completion network is trained in a weakly self-supervised way. We evaluate the depth prediction in both familiar and unknown environments. The results show that the predicted depth map is increasingly accurate and fine-grained as the density of the input LiDAR increases and our model can automatically choose the accumulation time of LiDAR data to adaptively balance the accuracy and speed. We also conduct complete autonomous landing experiments in a variety of real scenarios, verifying the effectiveness and robustness of our overall landing strategy. The videos of the real flight test are available at \url{https://youtu.be/0uj9LxWyMDA}.


\section{System Overview}

In this section, we introduce each module in the hardware system as well as the wire connection and software communication between them, illustrated in Fig.~\ref{wire}. The images of the UAV and sensors can be viewed in the supplementary video.

\subsection{UAV Platform}

The flight platform we used in this paper is DJI Matrice 600 Pro, a commercial hexacopter UAV equipped with a carbon-fiber airframe, the A3 Pro flight controller, Lightbridge 2 HD transmission system, six 4500mAh intelligent batteries, and a built-in battery management system. The vehicle is 1133mm in the diagonal wheelbase and weighs 9.5 kg including batteries. Its maximum payload capacity is about 6 kg. The flight time in hovering is about 16 minutes with full payload and 32 minutes without payload. The A3 Pro Flight Controller consists of a flight controller, three GPS-Compass Pro, two IMU Pro, and a Power Management Unit (PMU), which provides triple modular redundancy, improving the system's anti-risk performance. Self-adaptive systems will automatically adjust flight parameters based on different payloads. Other sensors and devices are mounted on our own designed two-layer carbon-fiber board at the bottom of the center airframe.

\begin{figure}[t]
\begin{center}
\includegraphics[width=\linewidth]{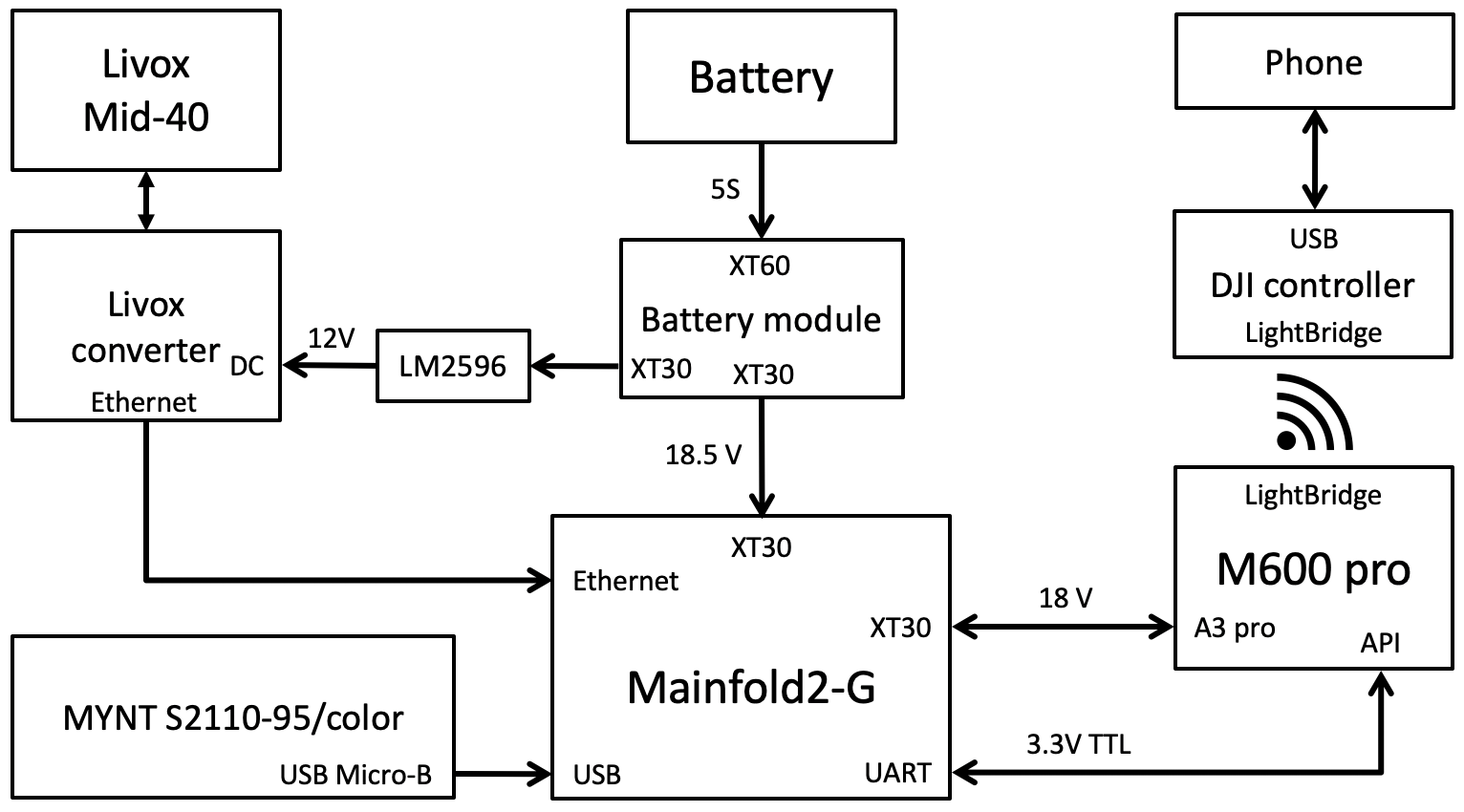}
\end{center}
\caption{UAV hardware system overview.}
\label{wire}
\end{figure}

\subsection{LiDAR}

We adopt Livox Mid-40 as our onboard LiDAR because it is lightweight, cost-effective, and has several unique advantages fit for UAV landing scenes. The Livox Mid-40 sensor weighs 760g which can be easily adapted to our flight platform. It costs only \$599, much cheaper than multi-line surround-view LiDAR. Its major advantage is to detect towards one side, thus very suitable for mount in the down-headed direction on the UAV.
Another advantage is that it features non-repetitive scanning patterns in which the areas scanned inside the FOV grow the longer the integration time, increasing the likelihood of objects and other details within the FOV being detected. It is equivalent to the 32-line product when the integration time is 0.1s. With an integration time of 0.5s, the coverage performance is equivalent to the 64-line product. As integration time increases, the FOV coverage will approach 100\%. This feature enables UAV to dynamically choose different LiDAR integration time according to the needs of the algorithm. Even without other sensors, only relying on LiDAR can ensure reliable depth perception.

\subsection{Stereo Camera}

In this paper, although we use the fusion of monocular image and LiDAR points for depth completion, the binocular camera is still necessary. It can provide more self-supervision during the training phase, and more importantly, helps to evaluate the accuracy of the predicted depth map during the inference phase. 
For this reason, we chose MYNT EYE Standard/Color (S2110-95/Color) binocular camera, which provides a stereo color image with a resolution of 1280*400@60FPS. The baseline length is 8cm and the FOV angle is $95^{\circ}$. To ensure the consistent quality of stereo images, the camera provides automatic ISP exposure and automatic white balance functions. The global shutter function also effectively reduces the image distortion during rapid shooting scenes. At the same time, the camera provides hardware-level time synchronization and binocular frame synchronization at millisecond-level accuracy.

\subsection{Onboard Computer}

To accelerate the inference speed of the depth completion model with modern GPU, we choose DJI Manifold 2-G as the onboard computer. Its processor is the NVIDIA Jetson TX2, which is built around an NVIDIA GPU with 256 CUDA cores and it has 8 GB 128-bit LPDDR4 memory. The added UART ports are used to connect with the A3 Pro flight controller for accessing the flight status and manipulating the UAV. Since Manifold 2-G supports two external power supply independently and can automatically choose the power source with a higher voltage, we use both a reserved output 18V power supply from the Matrice 600 Pro and a separate 5300mA 5S LiPo battery to power the computer. In this way, Manifold 2-G will prefer the separate battery because it has a higher voltage. This not only saves more power to the vehicle but also ensures the availability of the onboard computer when the separate battery fails. In the whole system, all sensors except LiDAR support global time synchronization, so we use Precision Time Protocol (PTP) protocol to synchronize the timestamp of the onboard computer and LiDAR.

\subsection{Others}

The DJI Power Distribution Unit is connected to the separate battery to power the Manifold 2-G and LiDAR since its electromagnetic compatibility can reduce interference to the GNSS or Wi-Fi signal from the power supply. However, because the battery voltage is higher than the working requirement of LiDAR, we use a DC-DC voltage converter module (LM2596) to provide 12V power for LiDAR.

Through the LightBridge HD transmission system, we can monitor the status of UAV and the algorithm result of the onboard computer in real-time. Also, we develop a mobile app based on DJI UXSDK to switch the flight mode of UAV, \ie, the auto landing mode and the manual control mode.

The software has been developed in C++ and Python as ROS (Robot Operating System) package, using the well known OpenCV libraries to process the images and PyTorch for model inference.


\begin{figure}[t]
\begin{center}
\includegraphics[width=\linewidth]{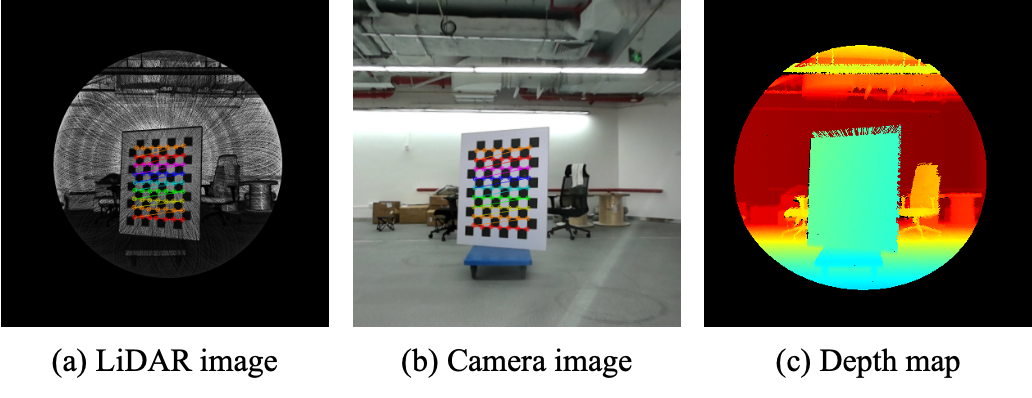}
\end{center}
\caption{(a)(b) Corner points in LiDAR and camera images; (c) Generated depth map corresponding to camera image.}
\label{calib}
\end{figure}

\section{Robust Landing Site Selection}

In this section, we first introduce the calibration between stereo cameras and LiDAR as well as their time and spatial alignment method for generating synchronized images and depth maps. Then we elaborate on the dynamic time depth completion model and its training and inference strategy. Finally, with the accurate depth estimation result, the overall landing site selection process is derived.

\subsection{Calibration and Alignment}

The calibration is the basis for dataset construction and depth completion algorithm. The intrinsic and extrinsic parameters of the stereo camera can be directly calculated using the existing calibration method~\cite{zhang2000flexible} implemented in OpenCV, while the calibration between LiDAR and the camera needs a special conversion. Compared to the traditional sparse multi-line LiDAR, Livox Mid-40 has the advantage of high FOV coverage. Therefore, we can directly visualize it as an image by projecting a few seconds of LiDAR points of a static scene to a virtual image plane and allocating the grayscale colors based on their intensity. In this way, the common chessboard calibration board can be used to extract the characteristic corner points from both images, as shown in Fig.~\ref{calib}. The corner points in the LiDAR image corresponds to the 3D points in the world while the corner points in the camera image are the corresponding 2D projections. This constructs a Perspective-n-Point (PnP) problem, which can be effectively resolved to acquire the rotation matrix $R_{c,l}$ and the translation $t_{c,l}$.

To generate the depth map corresponding to image $I_0$ captured at time $t_0$, we first perform time alignment to transform several seconds of LiDAR data to the LiDAR coordinate system $l_0$ at time $t_0$. 
Specifically, given a LiDAR point $p^{l_1}$ measured at time $t_1$, we can get $p^{l_0}$ by using affine transformation as $p^{l_0}=T_{l_0,l_1}p^{l_1}$, in which
$$T_{l_0,l_1}=T_{l_0,w}T_{w,l_1}=T_{l,i}T_{i_0,w}T_{w,i_1}T_{i,l}=T_{l,i}T_{w,i_0}^{-1}T_{w,i_1}T_{l,i}^{-1}$$
where $T_{l,i}$ is a fixed transformation matrix between LiDAR and IMU, and $T_{w,i_0}, T_{w,i_1}$ are the position and attitude of the IMU in the world coordinate system at time $t_0,t_1$.
However, since the data acquisition frequency of the various sensors is different, $T_{w,i_0},T_{w,i_1}$ are obtained by conducting a linear interpolation and a spherical linear interpolation to GPS and IMU data respectively.

The second step is to spatially align the LiDAR point to the camera. Given a LiDAR point $p^l=\left (x^l,y^l,z^l \right )^T$, its coordinate in the camera coordinate system is $p^c=\left ( x^c,y^c,z^c \right )^T=R_{c,l} \cdot p^l + t_{c,l}$, the corresponding projection coordinate in the image plane is $(x,y)=(f \cdot x^c/z^c+c_x,f \cdot y^c/z^c+c_y) $, where $f$ is the focal length in pixel unit and $(c_x, c_y)$ is the principal center point. Since $x, y$ are generally not the integer, we assign it to the nearest pixel $(u, v)$. If one pixel is allocated multiple times, we only select the point closest to the image in time. The generated depth map is shown in Fig.~\ref{calib}.

\subsection{Dynamic Time Depth Completion}

Depth Completion is a depth estimation problem with partial known ground truth. So we adopt an encoder-decoder network structure as \cite{ma2019self}, which processes the image and sparse depth inputs separately with late fusion. 
In the following, we will detail our training and inference strategies to achieve robust and accurate depth prediction.

\textbf{Training Losses}
Given an input image as well as its corresponding sparse depth $d$, network prediction $pred$, and the ground truth depth map $gt$, we define the training loss $C$ as a combination of four main terms,
$$C=\alpha_{d}C_d+\alpha_{r}C_r+\alpha_{p}C_p+\alpha_{s}C_s$$
where $C_d,C_r,C_p,C_s$ are supervised depth loss, depth ratio loss, photometric loss, smoothness loss respectively, and $\alpha_{d},\alpha_{r},\alpha_{p},\alpha_{s}$ are their loss weights. 

Specifically, the supervised depth loss $C_d$ penalizes the differences between the network output and the ground truth depth map on the set of pixels with known sparse depth, which helps the training for fast convergence, defined as 
$$C_d=\left \| \mathds{1}_{\{gt>0\}} \cdot (pred-gt) \right \|_2^2$$

In addition, we propose a new depth ratio loss specialized for our scenarios. During the landing process, we expect that the predicted depth map becomes increasingly fine-grained with the altitude going down. Therefore, the depth error with the same absolute value should attach more supervision at a lower altitude, which can be realized by penalizing the proportion of the difference between the predicted and the ground truth value, defined as
$$C_r=\left \| \mathds{1}_{\{gt>0\}} \cdot \frac{pred-gt}{gt} \right \|_1$$

It is worth noting that the sparse depth input $d$ can be used in place of the ground truth $gt$ for $C_d$ and $C_r$ when forming a self-supervised training.

The photometric loss $C_p$ is to indirectly supervise the predicted depth by comparing the image similarity between the current image and the reconstruction image inversely warped from the nearby frame. 
Suppose the current image is captured by left side of the stereo camera at time $t_i$ denoted as $I_{l, i}$, we choose the adjacent two frames on the same side and the image with the same timestamp captured by the other side of the stereo camera as the nearby frames denoted as $I_{l,i-1}, I_{l,i+1}, I_{r,i}$ respectively. The relative pose between stereo cameras is fixed while the method of calculating the transformation matrix between the adjacent frame with known flight position and attitude is the same as that between the LiDAR point, elaborated in Section~3.1. With relative poses that denoted as $T_{i-1,i}$, $T_{i+1,i}$, and $T_{r,l}$, the reconstruction images can be obtained as $I_{l,i-1}{}'$, $I_{l,i+1}{}'$, and $I_{r,i}{}'$ referred to \cite{godard2017unsupervised}. So the photometric loss is defined as 
$$C_p=\frac{1}{3}\sum_{s}\alpha \frac{1-SSIM(I_{s}{}',I_{s})}{2}+(1-\alpha )\left \| I_{s}{}'-I_{s} \right \|_1$$
where $s\in \{\{l,i-1\},\{l,i+1\},\{r,i\}\}$ and $\alpha=0.85$. We use simplified SSIM~\cite{wang2004image} with a $3\times3$ block filter.

The smoothness loss $C_s$ forces a neighboring constraint to encourage depth to be locally smooth, which is the basis to achieve the stable result of terrain conditions with depth derivatives. Here we penalize the first and second-order derivatives of the depth predictions as 
$$C_s=\left \| \nabla pred \right \|_1+\left \| \nabla^2pred \right \|_1$$

\begin{figure}[t]
\begin{center}
\includegraphics[width=\linewidth]{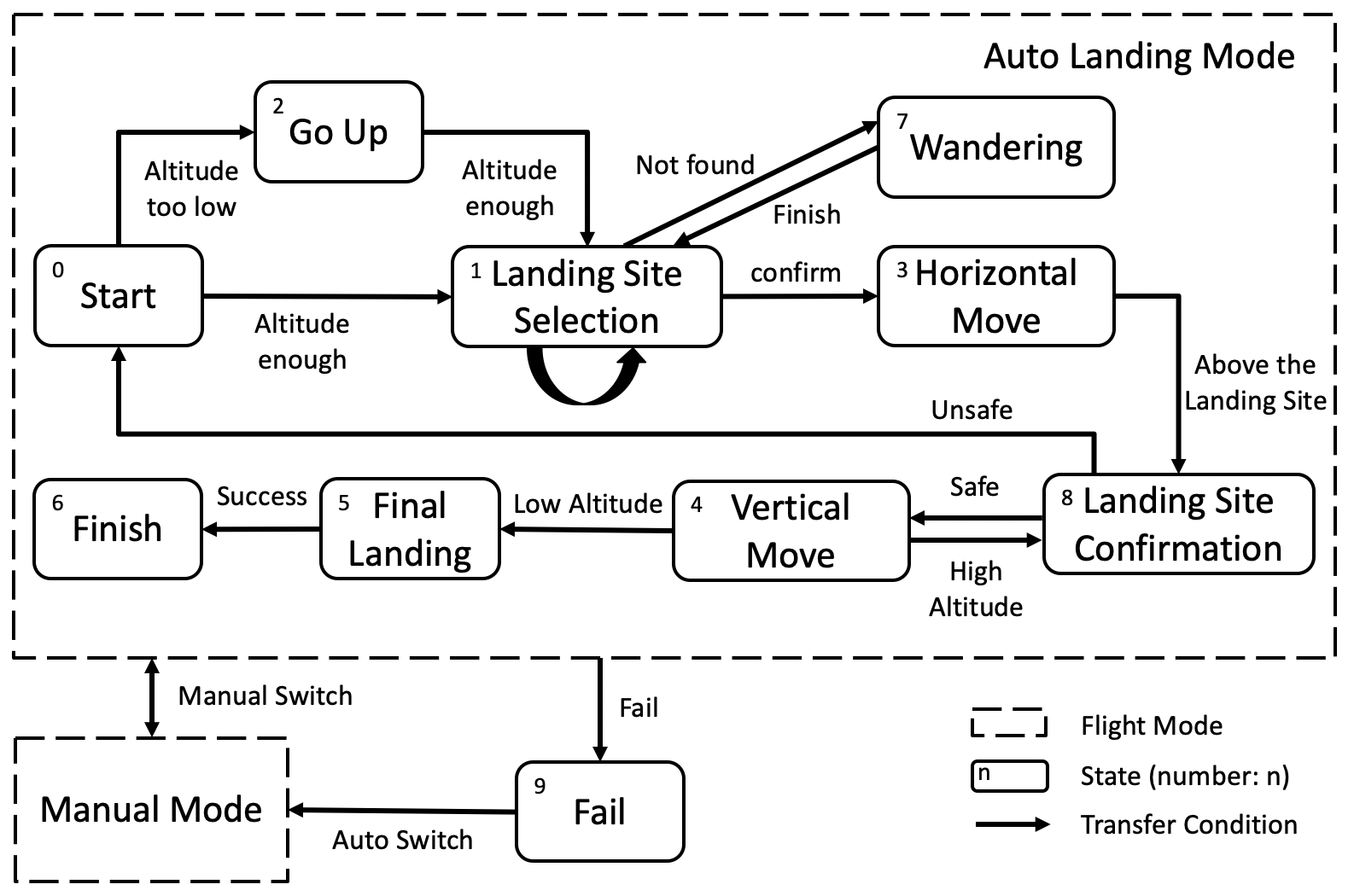}
\end{center}
\caption{Flow chart of autonomous landing.}
\label{flow}
\end{figure}

\textbf{Training Strategy}
We use a mixed training strategy that dynamically selects the density of the sparse depth for each batch. At 20\% of the time, the model is trained only with images where the input of the LiDAR branch is a zero tensor. In other situations, we randomly select 10\% to 50\% of LiDAR points to transform into a sparse depth map. 
This strategy improves the generalization ability at different inputs and realizes two useful characteristics that benefit our landing task. 1) It only requires a single image or LiDAR data as input, which maintains relatively stable performance even if one of the sensors fails. 2) The prediction result is gradually refined at the inference phase as the density of LiDAR points grows. This feature is very suitable for our LiDAR that has a non-repetitive scanning pattern. It ensures that given sufficient accumulation time of LiDAR, the predicted depth map will be accurate enough. 

\textbf{Self-evaluation Method}
Almost all existing depth completion algorithms are unable to self-assess and correct their prediction result at the inference stage. In our application, this may cause the UAV to select the wrong landing site and even crash. To improve the robustness of our model, we propose a self-evaluation method based on the natural character of the binocular camera. Similar to the principle of photometric loss, given the stereo images $I_l$, $I_r$, the predicted depth $pred$ according to $I_l$, the transformed sparse depth $d$, and the transformation matrix $T_{r,l}$ between stereo camera, we can generate the reconstructed right image $I_r^{pred}$ based on $I_l$, $pred$, and $T_{r,l}$. Without considering the imaging difference between two cameras caused by hardware and occlusion, if the predicted depth value is completely correct, the image similarity $sim_{pred}=SSIM(l_r^{pred}, I_r)$ will approach 1. Otherwise, $sim_{pred}$ is close to 0. Therefore, given $sim=SSIM(I_l, I_r)$ and $ssim_{d}=SSIM(I_r^{d}, I_r)$ as lower and upper bounds, we can evaluate $pred$ by comparing the relative size between $sim_{pred}$ and $sim$, $sim_{d}$. Conclude through the statistic, we find that when $sim_{pred}>(sim+sim_{d})/2$ or $sim_{pred}>sim+0.2$, the predicted depth is accurate enough for selecting the landing place.

\textbf{Inference Strategy}
Using this evaluation method, our dynamic time depth completion algorithm can be realized. In our system, the camera and the LiDAR produce the data at the frequency of 20hz and 10hz respectively. Every time we get an image, a coarse depth is first predicted. As the LiDAR data continues to acquire, the image and gradually denser depth input are combined to refine the output depth map. After each step, the prediction result is evaluated until the depth map is accurate enough. Generally, after about 1 second of accumulation of LiDAR, the coverage in the FOV is close to 100\%. If the result still doesn't meet the accuracy requirement, we directly perform the nearest neighbor interpolation on the sparse depth and set the depth outside the FOV area as unknown. 

\subsection{Landing Site Selection}

After obtaining an accurate depth map, we select the possible landing site through the following four steps.
1) We first convert the original `perspective' depth $pred$ to the `plane' depth $pred_p$ where all points in a plane parallel to the camera have the same depth. 
2) We then perform a perspective processing on $pred_p$ by correcting the roll and pitch to zero to obtain the depth map $pred_c$ that simulates the UAV is parallel to the ground.
3) The first and second-order derivatives of depth map $pred_c$ represents the slope and roughness of the ground. So we define that the safeness of ground should satisfies $\nabla pred_c < t_{inc}$ and $\nabla^2 pred_c < t_{tur}$, where $t_{inc}$ and $ t_{tur}$ are the maximum acceptable slope angle and roughness angle of the landing plane. According to this formula, we will calculate a binary mask $m$ that indicates the safeness of the ground. After performing a simple erosion and dilation to remove minor noises, a refined mask $m_r$ is generated.
4) Finally, we choose the largest inscribed circle of the safe area as the candidate landing place. The real radius $R$ of this circle can be calculated with the `plane' depth $p$ of the circle center and the radius $r$ in the image by $R=p/f*r$, where $f$ is the focal length. Referred to the UAV size, if $R>2m$, we hold that the candidate area is big enough and the center of the circle is selected as the safe landing site.

The above procedures correspond to state 1 in Fig.~\ref{flow}. However, it is worth noting that when selecting the landing site for each frame, we will first determine whether the previous landing site is available or not to keep the result stable and smooth. To avoid misjudgment by a single wrong prediction, only the landing site that is confirmed by five consecutive frames will be chosen. On the contrary, if no landing site is confirmed for about 5 seconds, then it will shift to state 7 which performs a random wandering with a non-repetitive global path. State 8 in Fig.~\ref{flow} is to perform a confirmation of the landing site. The only difference to state 1 is that the depth map is generated completely from dense LiDAR data, which can help to identify extremely small details of the dangerous area, such as stones or road curb.


\section{Evaluations}

\begin{table}[t]
\caption{Performance of different density of LiDAR input. $\downarrow$ means less is better and $\uparrow$ means higher is better.}
\label{tab}
\begin{center}
\resizebox{\linewidth}{12.5mm}{
\begin{tabular}{@{}cccccc@{}}
\toprule
P(LiDAR) & RMSE (mm) $\downarrow$ & REL (mm) $\downarrow$ & $\delta<0.25 \uparrow$ & SSIM $\uparrow$ \\ \midrule
0  & 1292.8 & 107.0  & 93.6\%  & 0.688    \\
0.1  & 1117.2 & 94.9  & 94.1\%  & 0.701    \\
0.3  & 944.7 & 78.4  & 95.5\%  & 0.724    \\
0.5  & 910.5 & 76.3  & 95.7\%  & 0.739    \\
1.0  & 861.3 & 48.8  & 97.9\%  & 0.764    \\
\bottomrule
\end{tabular}
}
\end{center}
\end{table}

\begin{figure}[t]
\begin{center}
\includegraphics[width=\linewidth]{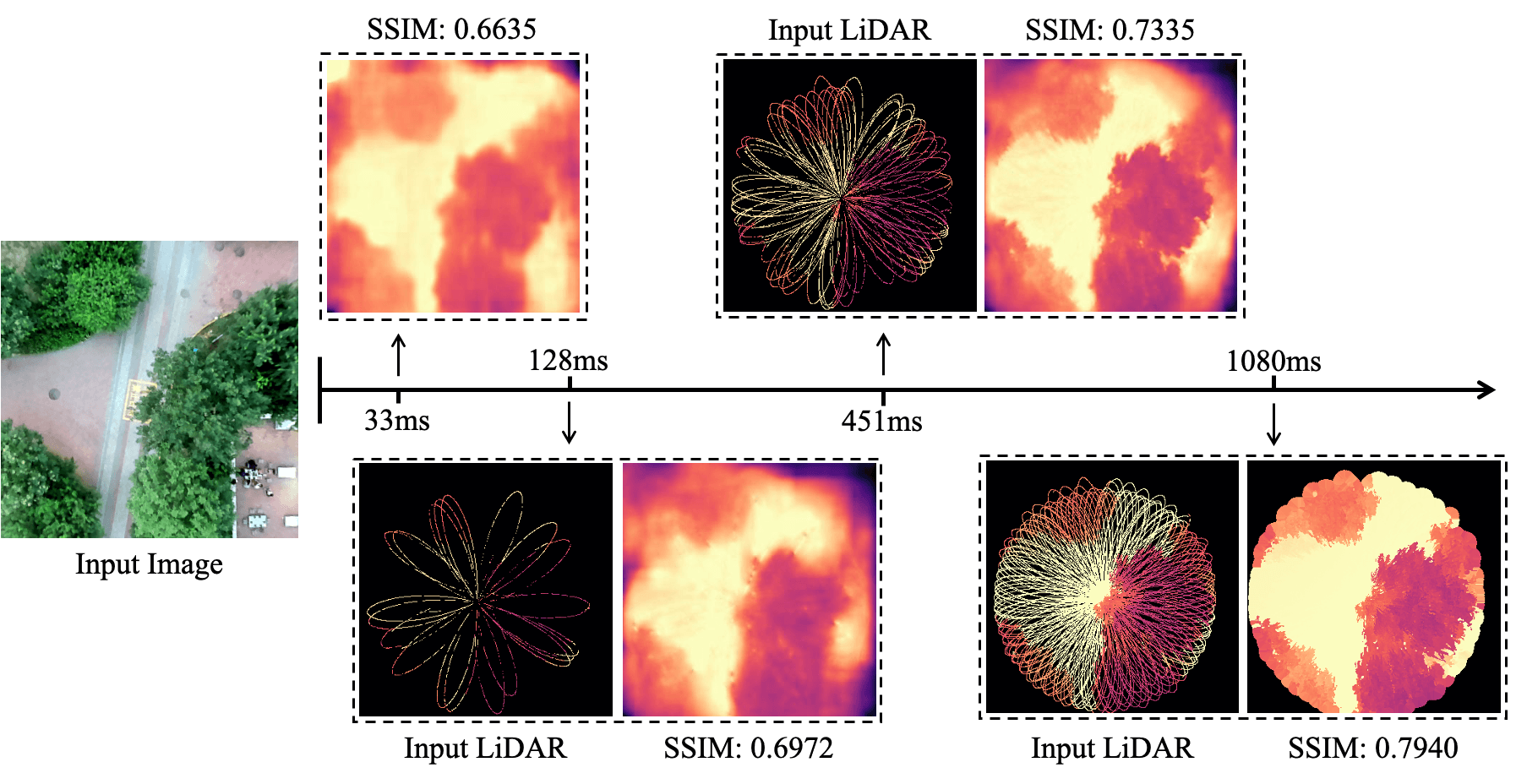}
\end{center}
\caption{Example timeline of predictions. As time progresses, the depth estimation becomes increasingly accurate.}
\label{timeline}
\end{figure}

\begin{figure}[t]
\begin{center}
\includegraphics[width=\linewidth]{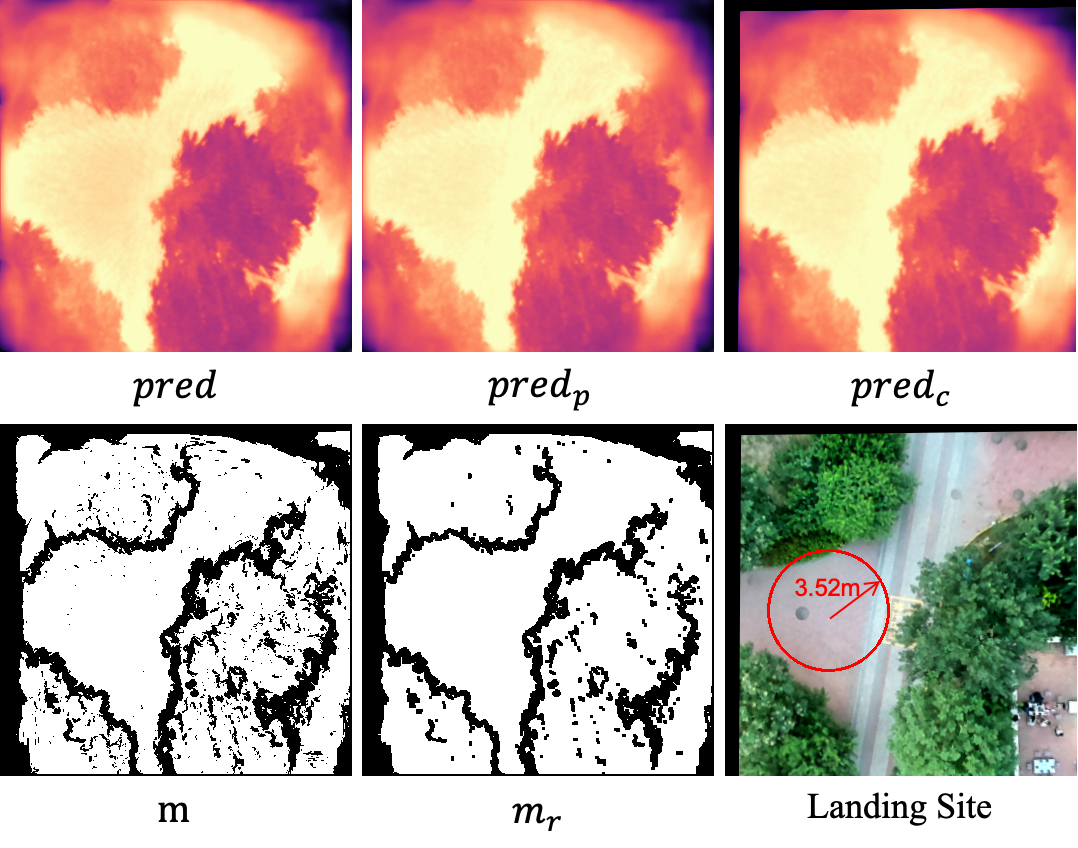}
\end{center}
\caption{Process of selecting the safe landing site.}
\label{circle}
\end{figure}

\subsection{Depth Completion Model}

We implement our depth completion network referred to \cite{ma2019self} with a backbone of ResNet-18~\cite{2016Deep}. To match the FOV range between the camera and the LiDAR, input images and sparse depth maps are center-cropped to a size of 352$\times$352. 
As mentioned in Section 3.2, a mixed training strategy by dynamically select the density of the input sparse depth is carried out to improve the generalization ability of the model.
We train the model for 30 epochs using Adam~\cite{kingma2014adam} optimizer with an initial learning rate of 2e-4 and a batch size of 16. After 20 epochs the learning rate is dropped to 2e-5 for better convergence. The loss weights are set to be $\alpha_{d}=1,\alpha_{r}=1,\alpha_{p}=2,\alpha_{s}=1$.

We collect 10 real flight records in different environments using onboard sensors. By performing time and spatial alignment described in Section~3.1 to collected data, we construct an aerial depth dataset consist of 30,000 sparsely annotated depth maps.
We then train our model on this dataset using both weak supervision of the sparse ground truth and the self-supervision of the image similarity. The average inference time of our model without any acceleration method running on NVIDIA Jetson TX2 is 33ms, which meet the real-time requirement.

We first conduct an experiment to compare the influence of the density of LiDAR input during the inference phase. The commonly used metrics such as RMSE, REL, and $\delta$ in the depth estimation problems~\cite{eigen2014depth} are used to evaluate the performance. We also introduce the average SSIM as a metric to indirectly evaluate the model through the quality of the reconstructed images. The numerical results under different test input choices are reported in Table.~\ref{tab} and a visualized timeline of the prediction is illustrated in Fig.~\ref{timeline}. 
In the beginning, we use a raw image without LiDAR data for prediction. The depth map can distinguish obvious altitude differences, such as trees and ground. As the density of the LiDAR increases, each metric is becoming better and the predicted depth map is more fine-grained. After about 1 second of accumulation, we directly interpolate the sparse depth map to get the most refined prediction map. This reduces the perception range of the UAV but makes it possible to observe a small height difference area such as grassland.

Next, we verify the effectiveness of the landing site selection strategy by setting $t_{inc}=10.0, t_{tur}=10.0$. It can be seen from Fig.~\ref{circle} that after depth conversion and perspective transformation, the safeness of the ground can be derived based on the accurate depth map $pred_c$ and the erosion and dilation can effectively avoid the interference of local abnormal points. 

\begin{figure}[t]
\begin{center}
\includegraphics[width=\linewidth]{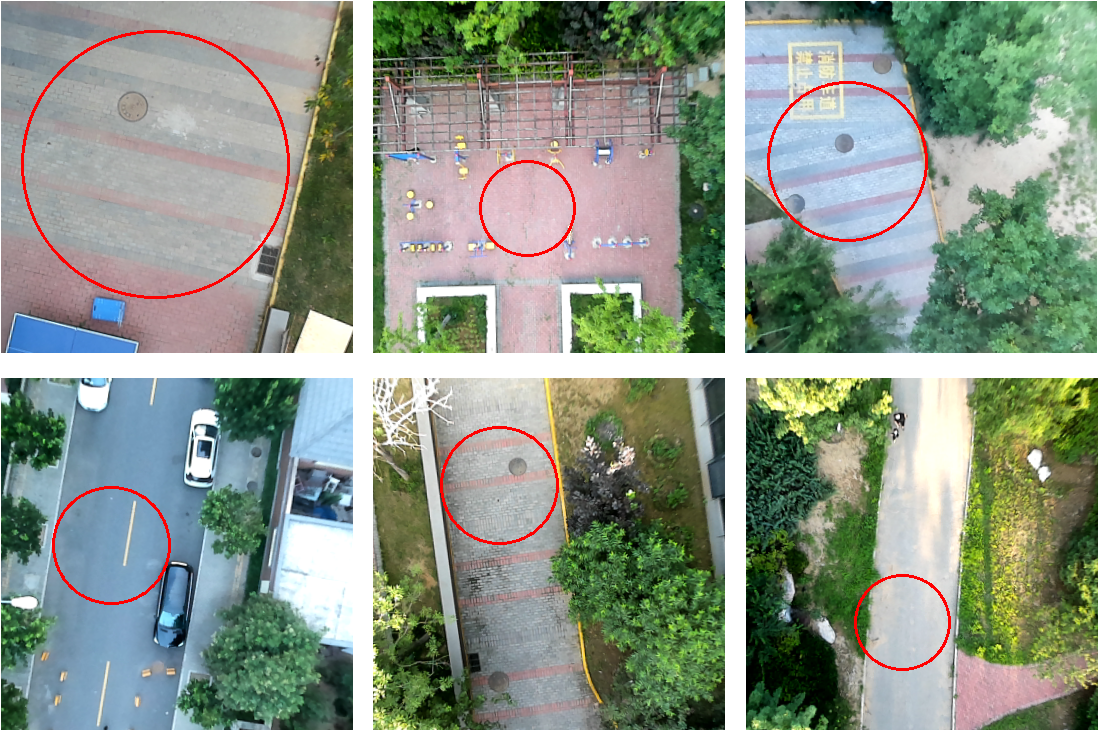}
\end{center}
\caption{Safe landing sites selected in real scenes.}
\label{site}
\end{figure}

\begin{figure}[t]
\begin{center}
\includegraphics[width=\linewidth]{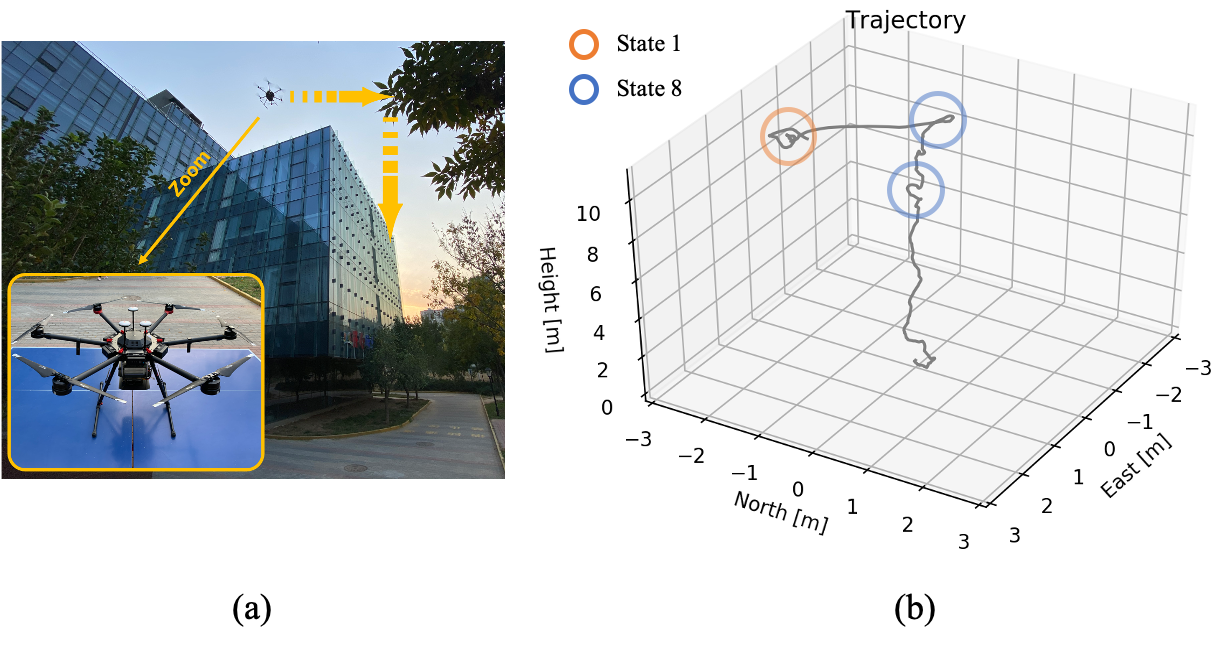}
\end{center}
\caption{(a) Our UAV system during the autonomous landing. (b) The corresponding 3D trajectory of the flight test in the local NED coordinate.}
\label{trajectory}
\end{figure}

\subsection{Real UAV Landing Experiments}

We conduct a series of autonomous landing experiments in a variety of environments where our UAV successfully finds the flat ground and land safely. At the same time, we record the average LiDAR accumulation time to obtain an accurate depth map for the depth completion model. In a familiar environment, the average time is 0.06 seconds, while in an unknown environment, this value reaches 0.7 seconds. This demonstrates that our model can adaptively balance the accuracy and speed to ensure the reliability of the prediction results. The typical safe landing sites selected in various real complex environments are shown in Fig.~\ref{site}. We also draw a landing trajectory in Fig.~\ref{trajectory}. It can be seen that the UAV performs a horizontal and gradually vertical movement in order after a few seconds of hovering to select a coarse landing site. The overall trajectory is smooth and almost no time is wasted on hovering except for the landing site selection and confirmation phase.


\section{Conclusions}

In this paper, we construct a UAV system to complete autonomous landing tasks in a non-cooperative environment by detecting the flat and safe ground area. Taking advantage of the characteristics of our perception system consists of low-cost LiDAR and stereo cameras, we come up with a dynamic time depth completion algorithm. Through the proposed self-evaluation method, it can dynamically select the LiDAR accumulation time to ensure an accurate depth prediction map. Through the real flight experiments, we verify that the model can adaptively balance the accuracy and speed, and the UAV can robustly select the safe landing sites even in a completely unknown environment. In future work, we consider expanding the recognition and tracking of dynamic objects into the system.



\bibliographystyle{IEEEtran}

\end{document}